\documentclass[10pt,twocolumn,letterpaper]{article}

\usepackage{wacv}
\usepackage{times}
\usepackage{epsfig}
\usepackage{graphicx}
\usepackage{amsmath}
\usepackage{amssymb}
\usepackage[ruled,vlined,linesnumbered]{algorithm2e}
\usepackage{booktabs}
\usepackage{caption}
\usepackage{subcaption}
\usepackage{pgfplots}
\pgfplotsset{compat=1.16}
\usepackage{comment}
\usepackage{array, makecell} 
\usepackage[numbers]{natbib}
\usepackage{xcolor}

\def\wacvPaperID{668} 

\wacvfinalcopy 

\ifwacvfinal
\def\assignedStartPage{1} 
\fi

\ifwacvfinal
\usepackage[breaklinks=true,bookmarks=false]{hyperref}
\else
\usepackage[pagebackref=true,breaklinks=true,colorlinks,bookmarks=false]{hyperref}
\fi

\ifwacvfinal
\else
\fi

\usepackage{amsmath,amsfonts,bm}

\def\eqref#1{equation~\ref{#1}}

\def\1{\bm{1}}

\DeclareMathAlphabet{\mathsfit}{\encodingdefault}{\sfdefault}{m}{sl}
\SetMathAlphabet{\mathsfit}{bold}{\encodingdefault}{\sfdefault}{bx}{n}

\usepackage{authblk}
\makeatletter
\renewcommand\AB@affilsepx{, \protect\Affilfont}
\makeatother

\begin{document}

\title{Affinity LCFCN: Learning to Segment Fish with Weak Supervision}

\author[1,2,4]{Issam Laradji}
\author[3]{Alzayat Saleh}
\author[2]{Pau Rodriguez}
\author[4]{Derek Nowrouzezahrai}
\author[3]{Mostafa Rahimi Azghadi}
\author[2]{David Vazquez}

\affil[1]{issam.laradji@gmail.com}
\affil[2]{Element AI}
\affil[3]{James Cook University}
\affil[4]{McGill University}

\maketitle

\begin{abstract}
Aquaculture industries rely on the availability of accurate fish body measurements, e.g., length, width and mass. Manual methods that rely on physical tools like rulers are time and labour intensive. Leading automatic approaches rely on fully-supervised segmentation models to acquire these measurements but these require collecting per-pixel labels -- also time consuming and laborious: i.e., it can take up to two minutes per fish to generate accurate segmentation labels, almost always requiring at least some manual intervention.  We propose an automatic segmentation model efficiently trained on images labeled with only point-level supervision, where each fish is annotated with a single click. This labeling process requires significantly less manual intervention, averaging roughly one second per fish. Our approach uses a fully convolutional neural network with one branch that outputs per-pixel scores and another that outputs an affinity matrix. We aggregate these two outputs using a random walk to obtain the final, refined per-pixel segmentation output. We train the entire model end-to-end with an LCFCN loss, resulting in our A-LCFCN method. We validate our model on the DeepFish dataset, which contains many fish habitats from the north-eastern Australian region. Our experimental results confirm that A-LCFCN outperforms a fully-supervised segmentation model at fixed annotation budget. Moreover, we show that A-LCFCN achieves better segmentation results than LCFCN and a standard baseline. We have released the code at \url{https://github.com/IssamLaradji/affinity_lcfcn}.
\end{abstract}

\section{Introduction}
Fish habitat monitoring is an important step for sustainable fisheries, as we acquire important fish measurements such as size, shape and weight. These measurements can be used to judge the growth of the fish and act as reference for feeding, fishing and conservation~\cite{ying2000application}. Thus, it helps us identify which areas require preservation in order to maintain healthy fish stocks. 

\begin{figure}[ht]
\centering
\includegraphics[width=\columnwidth]{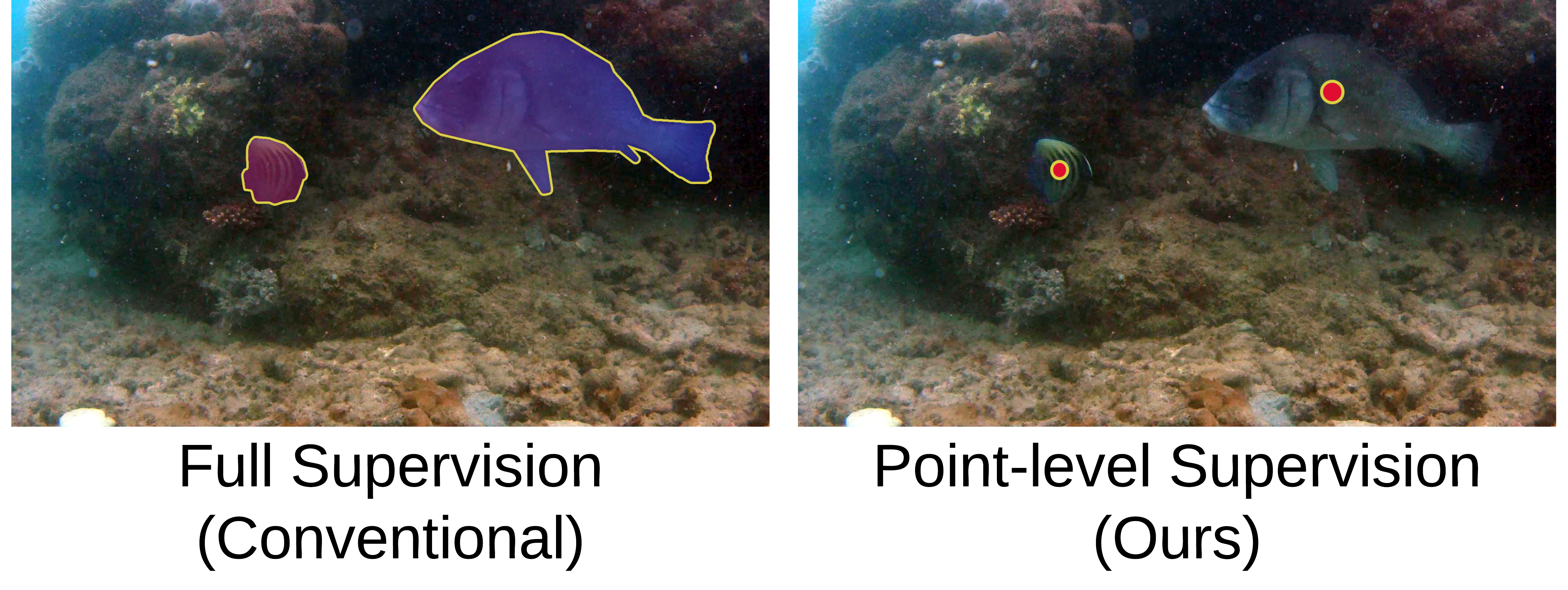}
\caption{\textbf{Labeling Scheme.}  Point-level supervision places a single  point on each fish body, whereas full supervision provides the full masks.}
\label{fig:Labeling}
\end{figure}

The UN Food and Agriculture Organization found that 33 percent of commercially important marine fish stocks worldwide are over-fished~\cite{delgado2003fish}. This finding is attributed to the fact that fishing equipments often catch unwanted fish that are not of the right size~\cite{roda2019third}. Catching unwanted fish can lead to more time needed to sort them. It  can also lead to more fuel consumption as these fish are extra weight on the boat, and cause long-term negative impact on the fisheries~\cite{garcia2019automatic}. Thus, acquiring fish size information has many important applications.
 
Many methods for measuring fish size are based on manual labor. Some experienced fishers are able to estimate length by eye. Other fishers use a ruler to measure the length~\cite{strachan1993length}. More recently, fishermen use echosounders to get the fish size but these tools are still on trail~\cite{pobitzer2015pre, berges2017practical}. Unfortunately these methods are time consuming, labour intensive and can cause significant stress to the fish~\cite{beddow1996predicting, booman1997efficiency}. 

Therefore, image segmentation systems for fish analysis~\cite{yu2020segmentation, garcia2020automatic, hao2015measurement} have gained lots of traction within the research community due to their potential efficiency. They can be used to segment fishes in an image in order to acquire morphological measurements such as size and shape. These systems can be installed in a trawl or underwater to cluster fish based on their sizes~\cite{garcia2019automatic}. Promising methods for image segmentation are based on deep learning, such as fully Convolutional Neural Networks (CNN) which now dominate many computer vision related fields. FCN8~\cite{long2015fully} and ResNet38D~\cite{wu2019wider} have shown to achieve promising performance in several segmentation tasks. In this work, we use a segmentation network based on FCN8 with an ImageNet~\cite{ILSVRC15} pretrained VGG16~\cite{Simonyan2014VGG} backbone.

Most segmentation algorithms are fully supervised~\cite{long2015fully, chen2018deeplab, jegou2017one}, as they require per-pixel annotations in order to train. These annotations are prohibitively expensive to gather due to the requirement of field expert annotators, a specialized tool, and intensive labor. In order to reduce these annotation costs, weakly supervised methods were proposed to leverage annotations that are cheaper to acquire. The most common labeling scheme is image-level annotation~\cite{rania2018a, ahn2019a}, which only requires a global label per image. Other forms of weak supervision are scribbles~\cite{vernaza2017a} and bounding boxes~\cite{Hu_2018} which were shown to improve the ratio of labeling effort to segmentation performance. In this work, we use point-level annotations since they require a similar acquisition time as image-level annotations, while significantly boosting the segmentation performance~\cite{bearman2016s}. Unfortunately, methods that use point-level supervision either need training a proposal network~\cite{laradji2019instance} or tend to output large blobs that do not conform to the segmentation boundaries~\cite{bearman2016s}. Thus, these methods are not well suited to images with objects of specific boundaries like fish. A promising weakly supervised method is LCFCN~\cite{laradji2018blobs}, which is better at localizing multiple objects but  does not segment the objects correctly. In this work we build on LCFCN to improve its segmentation capabilities.

\citet{ahn2018learning} showed that it is possible to train a segmentation network with image-level annotations by learning to predict a pixel-wise affinity matrix. This matrix is a weighted graph where each edge represents the similarity between each pair of pixels~\citep{shi2000normalized, levin2007closed}. However, in \citet{ahn2018a} the process to obtain this affinity matrix is costly and depends heavily on proxy methods such as Class Activation Map (CAM)~\citep{ahn2018a} to approximate it. Given the advantages of affinity networks for image segmentation, we propose a novel affinity module that automatically infers affinity weights. This module can be integrated on any standard segmentation network and it eliminates the need for explicit supervision such as acquiring pairs between pixels of CRF-refined CAMs~\citep{ahn2018a}. 

Therefore, we extend LCFCN with an affinity-based module in order to improve the output segmentation of the fish boundaries. Our model follows three main steps. First, features are extracted using a pre-trained backbone like ResNet38. Then, an activation branch uses these features to produce pixel-wise class scores. From the same backbone features, the affinity branch infers pairwise affinity scores between the pixels. Finally, the affinity matrix is combined with the pixel-wise class scores using random walk~\cite{lovasz1993random} to produce a segmentation mask. The random walk encourages neighboring pixels to have similar probabilities based on their semantic similarities. As a result, the predicted segmentations are encouraged to take the shape of the fish. During training, these segmentations are compared against the point-level annotations using the LCFCN loss~\cite{laradji2018blobs}. This loss ensures that only one blob is output per object which is important when there are multiple fish in an image. Unlike AffinityNet~\cite{ahn2018learning} which requires expensive pre-processing and stage-wise learning, the whole model can be trained end-to-end efficiently. Finally, the segmentation output by our model can be used to generate pseudo ground-truth labels for the training images. Thus, we can train a fully supervised network on these pseudo ground-truth masks achieving better results. The reason behind the improvement can be attributed to the fact that these networks can be robust against noisy labels~\cite{laradji2019masks}.

\begin{figure*}[htbp]
\centering
\includegraphics[width=\textwidth]{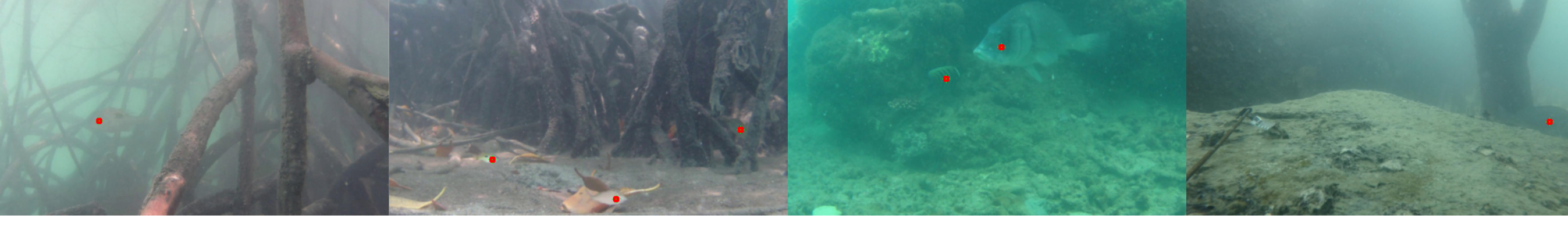}
\caption{\textbf{DeepFish Dataset.} Images from different  habitats with point annotations  on the fish (shown as red dots).}
\label{fig:Habitats}
\end{figure*}

We benchmark A-LCFCN on the segmentation subset of the {\it DeepFish}~\cite{saleh2020realistic} dataset. This dataset contains images from several habitats from north-eastern Australia (see Figure~\ref{fig:Habitats} for examples). These habitats represent nearly the entire range of coastal and nearshore benthic habitats frequently accessible to fish species in that area. Each image in the dataset has a corresponding segmentation label, where pixels are labelled to differentiate between fish pixels and background pixels (see Figure~\ref{fig:qualitative}). Our method achieved an mIoU of 0.879 on {\it DeepFish}~\cite{saleh2020realistic}, which is significantly higher than standard point-level supervision methods, and fully-supervised methods when the annotation budget is fixed.


For our contributions, (1) we propose a framework that can leverage point-level annotations and perform accurate segmentation of fish present in the wild. (2) We propose an affinity module that can be easily added to any segmentation method to make the predictions more aware of the segmentation boundaries. (3) We present results that demonstrate that our methods achieve significant improvement in segmentation over baselines and fully supervised methods when the annotation budget is fixed.

\section{Related Work}
\label{sec:related_work}
In this section, we first review methods applied to general semantic segmentation, followed by semantic segmentation for fish analysis. Then we discuss affinity methods that use pair-wise relationships between the pixels for improved segmentation. Finally, we discuss weakly supervised methods for segmentation and object localization.

\paragraph{Semantic Segmentation}
is an important computer vision task that can be applied to many real-life applications~\cite{long2015fully, chen2018deeplab, jegou2017one}. This task consists of classifying every object pixel into corresponding categories. Most methods are based on fully convolutional networks which can take an image of arbitrary size and produce a segmentation map of the same size. Methods based on Deeplab~\cite{chen2018deeplab} consistently achieve state-of-the-art results as they take advantage of dilated convolutions, skip connections, and Atrous Spatial Pyramid Pooling (ASPP) for capturing objects and image context at multiple scales. However, these methods require per-pixel labels in order to train, which can result in expensive human annotation cost when acquiring a training set for a semantic segmentation task.

\paragraph{Semantic Segmentation Methods for Fish Analysis}
have been used for efficient, automatic extraction of fish body measurements~\cite{Fernandes2020DeepTilapia}, and prediction of their body weight~\cite{Fernandes2020DeepTilapia, konovalov2018estimating, konovalov2019automatic} and shape for the purposes of preserving marine life. \citet{garcia2019automatic} used fully-supervised segmentation methods and the Mask R-CNN~\cite{he2017mask} architecture to localize and segment each individual fish in underwater images to obtain an estimate of the boundary of every fish in the image for estimating fish sizes to prevent catches of undersized fish. \citet{FrenchConvolutionalVideo} presented a fully-supervised computer vision system for segmenting the scenes and counting the fish from CCTV videos installed on fishing trawlers to monitor abandoned fish catch. While we also address the task of segmentation for fish analysis, to the best of our knowledge, we are the first to consider the problem setup of using point-level supervision, which can considerably lower the annotation cost.

\paragraph{Affinity-based Methods for Semantic Segmentation} have been proposed to leverage the inherent structure of images to improve segmentation outputs~\cite{krahenbuhl2011efficient, liu2015semantic, chen2015learning}. They consider the relationship between pixels which naturally have strong correlations. Many segmentation methods use conditional random fields (CRF)~\cite{chen2017deeplab, krahenbuhl2011efficient} to post-process the final output results. The idea is to encourage pixels that have strong spatial and feature relationships to have the same label. CRF were also incorporated to a neural network as a differentiable module to train jointly with the segmentation task~\cite{liu2015semantic}. Others leverage image cues based on grouping affinity and contour to model the image structure~\cite{maire2016affinity, liu2017learning}. Most related to our work is \citet{ahn2018learning} which proposes an affinity network that learns from pairwise samples of pixels labeled with a segmentation network and a CRF. The network is then used to output an affinity matrix which is used to refine the final segmentation output. Unfortunately, these methods require expensive iterative inference procedures, and require to learn the segmentation task in stages. In our work, we use part of the affinity network as a module that can be incorporated to any segmentation network, adding minimal computational overhead while increasing the model's sensitivity to object boundaries and segmentation accuracy.

\paragraph{Weakly Supervised Semantic Segmentation} methods have risen in popularity due to their potential in decreasing the human cost in acquiring a training set. \citet{bearman2016s} is one of the first methods that use point-supervision to perform semantic segmentation. They showed that manually collecting image-level and point-level labels for the PASCAL VOC dataset~\citep{everingham2010pascal} takes only $20.0$ and $22.1$ seconds per image, respectively. This scheme is an order of magnitude faster than acquiring full segmentation labels, which is $239.0$ seconds. The most common weak supervision setup is using image-level labels to perform segmentation~\cite{ rania2018a, ahn2019a}. They use a wide range of techniques that include affinity learning, self-supervision, and co-segmentation. However, these methods were applied to the PASCAL VOC~\cite{everingham2010pascal} dataset that often has large objects. In our work we consider underwater fish segmentation with point-level supervision which has its own unique challenges.

\paragraph{Weakly Supervised Object Localization} methods can be an important step for segmentation as they allow us to identify the locations of the objects before grouping the pixels for segmentation. \citet{ren2015faster, yolov3} are current state-of-the-art methods for object localization, but they require bounding boxes. However, several methods exist that use weaker supervision to identify object locations~\cite{song2014learning, song2014weakly, lempitsky2010learning, li2018csrnet, laradji2020looc, laradji2019masks, laradji2020weaklyWS, laradji2020weaklyAL}. Close to our work is LCFCN~\cite{laradji2018blobs} which uses point-level annotations in order to obtain the locations and counts of the objects of interest. While this method produces accurate counts and identifies a partial mask for each instance, it does not produce accurate segmentation of the instances. Thus, we extend this method by using an affinity-based module that takes pairwise pixel relationships into context in order to output blobs that are more sensitive to the object boundaries.

\section{Methodology}
\label{sec:methodology}

\begin{figure*}[!t]
\centering
\includegraphics[width=1.0\textwidth]{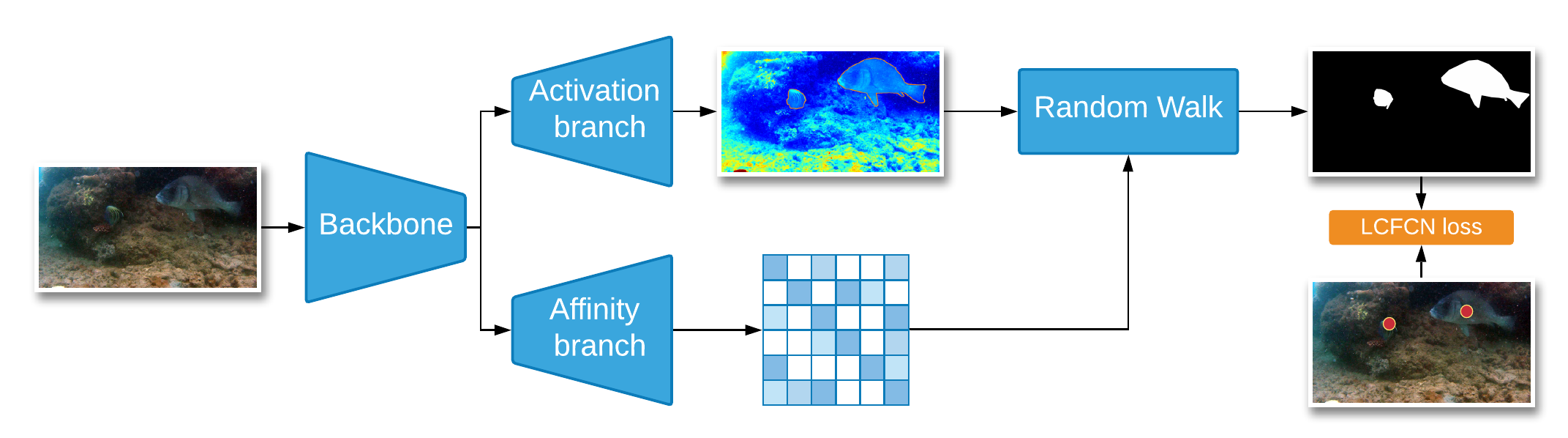}
\caption{\textbf{Affinity-based architecture.} 
The first component is the ResNet-38 backbone which is used to extract features from the input image. 
The second component is the activation branch outputs per-pixel scores. The third component is the affinity branch outputs an affinity matrix.
These two outputs are aggregated using a random walk to get the final, refined per-pixel output.}
\label{fig:model}
\end{figure*}

We propose A-LCFCN, which extends a fully convolutional neural network with an affinity-based module that is trained using the LCFCN loss. We consider the following problem setup. We are given $X$ as a set of $n$ training images with their corresponding set of ground-truth labels $Y$. $Y_i$ is a binary matrix of the same height $H$ and width $W$ as $X$ with non-zero entries that indicate the locations of the object instances. As shown in Figure~\ref{fig:Labeling}, there is a single non-zero entry per fish which is represented as a dot on top of the fish.

Shown in Figure~\ref{fig:model}, our model consists of a backbone $F^{bb}_\theta()$, an activation branch $F^{act}_\theta()$ and an affinity branch $F^{aff}_\theta()$. The backbone is a fully-convolutional neural network that takes as input an image of size $W \times H$ and extracts a downsampled feature map $f$ for the image. The activation branch takes the feature map as input and applies a set of convolutional and upsampling layers to obtain a per-pixel output $f^{act}$ as a heatmap that represents the spatial likelihood of the objects of interest. The affinity branch takes the same feature map as input and outputs a class-agnostic affinity matrix $f^{aff}$ that represents the pairwise relationships between the pixels. The affinity map and the activation map are then combined using random walk to refine the per-pixel output $f^{ref}$. This refinement adapts the output to be  aware of the semantic boundaries of the objects, leading to better segmentation. These components are trained collectively, end-to-end, using the LCFCN loss $\mathcal{L}_L$, which encourages each object to have a single blob. To further improve the performance, the trained model is used to output pseudo ground truth masks for the training images. These masks are then used as ground truth for training a fully-supervised network that is then validated on the test set. The details of this pipeline are laid out below.

\subsection{Obtaining the Activation Map}
The activation branch $F^{act}_\theta$ transforms the features $f$ obtained from the backbone to per-pixel class scores, and upsamples them to the size of the input image. 

\subsection{Obtaining the Affinity Matrix}
The affinity branch is based on the AffinityNet structure described in \citet{ahn2018learning}, and the goal is to predict class-agnostic semantic affinity between adjacent coordinate pairs on a given image. These affinities are used to propagate the per-pixel scores from the activation branch to nearby areas of the same semantic object to improve the segmentation quality.

The affinity branch outputs a convolutional feature map $f^{aff}$ where the semantic affinity between a pair of feature vectors is defined in terms of their L1 distance as follows,
\begin{equation}
    W_{ij} = \exp \{ - ||f^{\text{aff}}(x_i, y_i) -  f^{\text{aff}}(x_j, y_j)||_1\},
    \label{eq:affinity}
\end{equation}
where $(x_i, y_i)$ indicates the coordinate of the $ith$ feature on feature map $f^{aff}$. 

In contrast to AffinityNet~\cite{ahn2018learning}, we do not require affinity labels for feature pairs to train our affinity layers. These layers are directly trained using the LCFCN loss on the point-level annotations as described in Section~\ref{sec:lcfcn}. 

\subsection{Refining the Activation Map with Affinity}
The affinity matrix is used to refine the activation map to diffuse the per-pixel scores within the object boundaries. As explained in \citet{ahn2018learning}, the affinity matrix is first converted to a transition probability matrix by first applying the Hadamard power on $W$ with value $\beta$ to get $W^{\beta}$ and normalizing it with row-wise sum on $W^{\beta}$. This operation results in the following transition matrix:
\begin{eqnarray}
    T = D^{-1} W^{\beta}, \ \ \textrm{where} \ \ D_{i i} = \sum_j W_{i j}^{\beta}. \label{eq:trans_mat}
\end{eqnarray}
higher $\beta$ makes the affinity propagation more conservative as it becomes more robust against small changes in the pairwise distances in the feature space.

Using the random walk described in \citet{ahn2018learning} we perform matrix multiplication of $T$ on the activation map $f^{act}$ for $t$ iterations to get the refined activations $f^{ref}$.

\subsection{Training the Weakly Supervised Model}
\label{sec:lcfcn}
The goal of our training strategy is to learn to output a single blob per fish in the image using point-level annotations (Figure~\ref{fig:Labeling}). Thus we use the LCFCN loss described in \citet{laradji2018blobs} as it only requires point-level supervision. While this was originally designed for counting, it is able to locate objects and segment them. On the refined activation output $f^{ref}$, we obtain per-pixel probabilities by applying the softmax operation to get $S$ which contains the likelihood that a pixel either belongs to the background or fish. The LCFCN loss $\mathcal{L}_L$ is then defined as follows:
\begin{equation}
\begin{split}
\mathcal{L}_L  = \underbrace{\mathcal{L}_I(S,Y)}_{\text{Image-level loss}} + \underbrace{\mathcal{L}_P(S,Y)}_{\text{Point-level loss}} + \underbrace{\mathcal{L}_S(S,Y)}_{\text{Split-level loss}} + \underbrace{\mathcal{L}_F(S,Y)}_{\text{False positive loss}},
\label{eq:loss_lcfcn}
\end{split}
\end{equation}
where $T$ represents the point annotation ground-truth. It consists of an image-level loss ($\mathcal{L}_I$) that trains the model to predict whether there is an object in the image; a point-level loss ($\mathcal{L}_P$) that encourages the model to predict a pixel for each object instance; a split-level ($\mathcal{L}_S$) and a false-positive ($\mathcal{L}_F$) loss that enforce the model to predict a single blob per instance (see~\cite{laradji2018blobs} for details for each of the loss components).

Applying the LCFCN loss on the original activation map usually leads to small blobs around the center of the objects which form poor segmentation masks. However, with the activation map refined using the affinity matrix, the predicted blobs make better segmentation of the located objects. We call our method A-LCFCN as an LCFCN model that uses an affinity-based module. We summarize its training procedure in Algorithm~\ref{algo1}.

\subsection{Training on Pseudo Ground-truth Masks}
A trained A-LCFCN can be used to output a refined activation map for each training image. These maps are used to generate pseudo ground-truth segmentation labels for the training images. The outputs are first upsampled to the resolution of the image by bilinear interpolation. For each pixel, the class label associated with the largest activation score is selected, which could be either background or foreground. This procedure gives us segmentation labels for the training images which can be used for training a fully-supervised segmentation network, which could be any model such as DeepLabV3~\cite{chen2017rethinking}. At test time, the trained fully-supervised segmentation network is used to get the final segmentation predictions.

\subsection{Network Architecture}
While our framework can use any fully convolutional architecture, we chose a ResNet38 model based on the version defined in \citet{ahn2018learning} due to its ability to recover fine shapes of objects. However, instead of having two networks, one for the affinity output and one for the activation output, we used a shared ResNet38 as the backbone which we found to improve the results and speed up training. 

The affinity branch consists of three layers of 1$\times$1 convolution with 64, 128, 256 channels, respectively, to be applied on 3 levels of feature maps from the backbone. The results are bilinearly upsampled to the same size and concatenated as a single feature map. This feature map then goes through a 1$\times$1 convolution with 448 channels to obtain affinity features.

The activation branch consists of one 1x1 convolution with 2 channels. It is applied on the last feature map of the backbone to obtain the background and the foreground activation map. These activation maps are refined using random walk with the affinity branch to get improved segmentations.

For the fully supervised segmentation model that is trained on the pseudo ground-truth masks, we use a model that consists of a backbone that extracts the image features and an upsampling path that aggregates and upscales feature maps to output a score for each pixel. The backbone is an ImageNet pretrained network such as ResNet38~\cite{ahn2018learning} and the upsampling layers are based on FCN8~\cite{long2015fully}. The output is a score for each pixel $i$ indicating the probability
that it belongs to background or foreground. The final output is an argmax between the scores to get the final segmentation labels.

\newcommand\mycommfont[1]{\footnotesize\ttfamily{#1}}
\SetCommentSty{mycommfont}

\begin{algorithm}
\caption{Model Training}
\label{algo1}
\DontPrintSemicolon
\SetAlgoLined
\SetKwInOut{Input}{Input}
\SetKwInOut{Output}{Output}
\SetKwInOut{Parameter}{Parameters}
\Input{$X$ images, $Y$ point-level masks.}
\Output{Trained parameters $\theta^*$}
\BlankLine

\For{each batch $B$} {
$\mathcal{L} \gets 0$\\
\For{each $(X_i, Y_i) \in B$} {
  \emph{Extract features from the backbone}\\
  $f_i \gets F^{bb}_\theta(X_i)$\\[0.1in]
  
  \emph{Obtain the activation map}\\
  $f^{act}_i \gets F^{act}_\theta(f_i)$\\[0.1in]
  
  \emph{Obtain the affinity map}\\
  $f^{aff}_i \gets F^{aff}_\theta(f_i)$\\[0.1in]
  
   \emph{Get the transition matrix}\\
  $T_i \gets \text{Apply Eq. (1) and (2) on } f^{aff}_i$\\[0.1in]
  
  \emph{Refine the activation map}\\
  $f^{ref}_i \gets RandomWalk(f^{act}_i, T_i)$\\[0.1in]
  
  \emph{Compute the LCFCN loss}\\
  $\mathcal{L} \gets \mathcal{L} + \mathcal{L}_{L}(f^{ref}_i, Y_i)$\\
}
Update $\theta$  by backpropagating w.r.t. $\mathcal{L}$
}
\end{algorithm}

\section{Experiments}
We evaluate our models on two splits of the DeepFish dataset~\cite{saleh2020realistic}, {\it FishSeg} and {\it FishLoc} to compare segmentation performance. We show that our method A-LCFCN outperforms the fully supervised segmentation method if the labeling effort between acquiring per-pixel labels and point annotations is fixed. Further, we show that our method outperforms other methods that do not use affinity. We further show that training on pseudo ground-truth masks generated by A-LCFCN using a fully segmentation model boosts segmentation performance even further. 

\subsection{DeepFish~\cite{saleh2020realistic}}
The {\it DeepFish} dataset\footnote{Found here: \url{https://github.com/alzayats/DeepFish}} consists of around 40 thousand images obtained from 20 different marine habitats in tropical Australia (Figure~\ref{fig:Habitats}). For each habitat, a fixed camera has been deployed underwater to capture a stream of images over a long period of time.  The purpose is to understand fish dynamics, monitor their count, and estimate their sizes and shapes.
 
The dataset is divided into 3 groups: {\it FishClf} that contains classification labels about whether an image has fish or not, {\it FishLoc} that contains point-level annotatons indicating the fish location, and {\it FishSeg} that contains segmentation labels of the fish. Since our models require at least point-level supervision, we use {\it FishLoc} and {\it FishSeg} for our benchmarks.

\paragraph{FishLoc Dataset.} It consists of 3200 images where each image is labeled with point-level annotations indicating the locations of the fish. It is divided into a training set (n = 1600), a validation set (n = 640), and a test set (n = 960). The point-level annotations are binary masks, in which the non-zero entries represent the (x, y) coordinates around the centroid of each fish within the images (Figure~\ref{fig:Habitats}).
 
\begin{figure*}[hbt!]
  \centering
  \includegraphics[trim={0 0.8cm 0 0},clip,width=\textwidth]{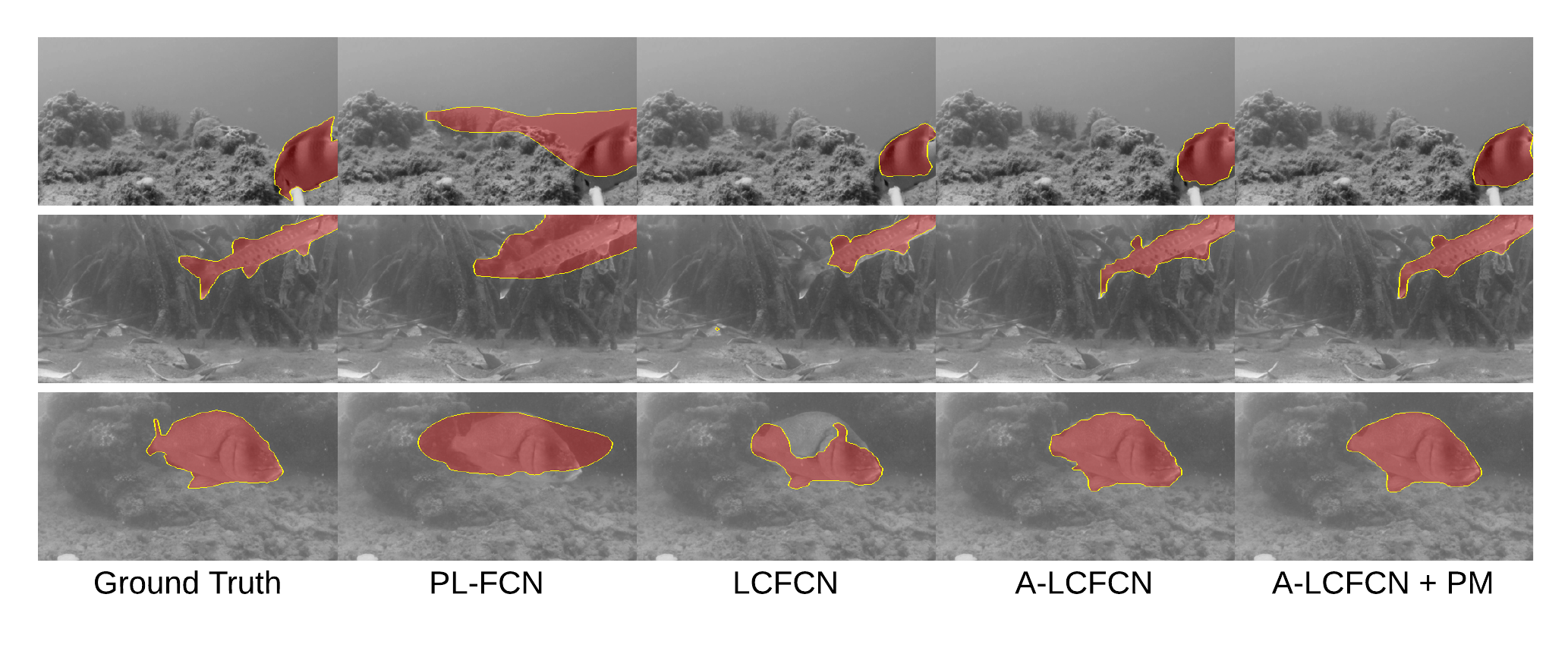}
  \caption{\textbf{Qualitative results.} We show the predictions obtained from training LCFCN and A-LCFCN. With the affinity branch the predictions are much closer to the ground-truth labels.}
  \label{fig:qualitative}
\end{figure*}

\paragraph{FishSeg Dataset.} It consists of 620 images with corresponding segmentation masks (see Figure\ref{fig:qualitative}),
separated into a training set (n = 310), validation set (n = 124), and a test set (n = 186). The images are resized into a fixed dimension  $256 \times 455$  pixels and normalized using ImageNet statistics~\cite{ILSVRC15}. According to \citet{saleh2020realistic}, it takes around 2 minutes to acquire the segmentation mask of a single fish. From the segmentation masks, we acquire point-level annotations by taking the pixel with the largest distance transform of the masks as the centroid (Figure~\ref{fig:Labeling}). These annotations allow us to train weakly supervised segmentation models.

Our models were trained either on FishLoc's or FishSeg's training set. For both cases we use FishSeg's test set to evaluate the segmentation performance. We have removed training images from FishLoc that overlap with FishSeg's test set for reliable results.

\subsection{Evaluation Procedure}
\label{sec:evaluation}
We evaluate our models against Intersection over Union (IoU), which is a standard metric for semantic segmentation that measures the overlap
between the prediction and the ground truth: $IoU = \frac{TP}{TP + FP + FN}$
where TP, FP, and FN is the number of true
positive, false positive and false negative pixels across all
images in the test set.

We also measure the model's efficacy in predicting the fish count using mean absolute error which is defined as, $ MAE=\frac{1}{N}\sum_{i=1}^N|\hat{C}_i-C_i|, $ where $C_i$ is the true fish count for image $i$ and $\hat{C}_i$ is the  model's predicted fish count for image $i$. This metric is standard for object counting~\cite{guerrero2015Trancos, lempitsky2010learning} and it measures the number of miscounts the model is making on average across the test images. 

We also measure localization performance using Grid Average Mean Absolute Error (GAME)~\cite{guerrero2015Trancos} which is defined as,$\operatorname{GAME}(L)=\frac{1}{N}  \sum_{i=1}^{N}\left(\sum_{l=1}^{4^{L}}\left|\hat{C}_i^{l}-c_{i}^{l}\right|\right),$ where, $\hat{C}_i^{l}$ is the estimated count in a region $l$ of image $n$, and $c_{i}^{l}$ is the ground truth for the same region in the same image. The higher $L$, the more restrictive the $GAME$ metric will be. 
We present results for $GAME(L=4)$ which divides the image using a grid of $256$ non-overlapping regions where we compute the sum of the MAE across these sub-regions.

\subsection{Methods and Baselines}
We compare our method against two other weakly supervised image segmentation methods and a fully-supervised method. All these methods use the same feature extracting backbone of ResNet38, which we describe below.

\paragraph{Fully supervised fully convolutional neural network (FS-FCN).} This method is trained with the true  per-pixel class labels (full supervision). It combines a  weighted cross-entropy loss and weighted IoU loss as defined in Eq.(3) and (5) from \citet{wei2019f3net}, respectively. It is an efficient method that can learn from ground truth segmentation masks that are imbalanced  between different classes. In our case thee number of pixels corresponding to fish is much lower than those to the background.

\paragraph{Point-level loss (PL-FCN).} This method uses the loss function described in \citet{bearman2016s} which minimizes the cross-entropy against the provided point-level annotations. It also encourages all pixel predictions to be background for background images. 

\paragraph{LCFCN.} This method is trained using the loss function proposed by \citet{laradji2018blobs} against point level annotations to produce a single blob per object and locate objects effectively. LCFCN is based on a semantic segmentation architecture that is similar to FCN~\cite{long2015fully}. 
Since it was originally designed for counting and localization, LCFCN optimizes a loss function that ensures that only a single small blob is predicted around the centre of each object. This prevents the model from predicting large blobs that merge several object instances.

\paragraph{A-LCFCN (ours).} This method extends LCFCN by adding an  affinity branch as described in Section~\ref{sec:methodology}. Inspired by AffinityNet~\cite{ahn2018learning}, this branch predicts class-agnostic semantic similarity between pairs of neighbouring coordinates. The predicted similarities are used in a random walk~\cite{lovasz1993random} as transition probabilities to refine the activation scores obtained from the activation branch.

\paragraph{A-LCFCN + PM (ours).} 
This method first uses the output of a trained A-LCFCN on the training set to obtain pseudo mask labels. Then an FS-FCN is trained on these pseudo masks and is used to output the final segmentation results.

\paragraph{Implementation Details}

Our methods use an Imagenet~\cite{ILSVRC15} pre-trained ResNet38~\cite{wu2019wider}. The models are trained with a batch size of 1 for 1000 epochs with ADAM~\cite{kingma2014adam} and learning rates of $10^{-4}$, $10^{-5}$ and $10^{-6}$.  We report the scores on the test set of {\it FishSeg} using the model with the learning rate that achieved the best validation score. We used early stopping with patience of 10 epochs.

\begin{table*}[hbt!]
\centering
\caption{Comparison between methods on the FishSeg test set. {\bf Foreground} is the IoU between the predicted fish segmentation and their ground-truth, and {\bf Background} is the IoU between the predicted background segmentation and its ground-truth.}
\label{against_baselines}
\begin{tabular}{lcccccc}
\toprule
  & \multicolumn{3}{c}{\textbf{FishLoc}} & \multicolumn{3}{c}{\textbf{FishSeg}} \\
  & \textbf{Background} & \textbf{Foreground} & \textbf{mIoU} & \textbf{Background} & \textbf{Foreground} & \textbf{mIoU} \\
\midrule
FS-FCN$^*$ &   0.992 &     0.663 &    0.827 &   0.992 &     0.663 &    0.827 \\ \hline
PL-FCN &  0.931 &     0.214 &    0.573 &  0.910 &  0.173 &     0.542 \\ 
A-LCFCN  & 0.993 &    0.727 &   0.860 &    0.993 &    0.713 &   0.853 \\
LCFCN    & 0.989 &    0.559 &   0.774 &    0.992 &    0.684 &   0.838 \\
LCFCN+PM & \textbf{0.994} &    \textbf{0.764} &   \textbf{0.879} &    \textbf{0.993} &    \textbf{0.730} &   \textbf{0.862} \\
\bottomrule
\end{tabular}

\end{table*}

\subsection{Comparison against Weak Supervision}
We train the proposed method and baselines on the {FishSeg} and {FishLoc} training sets and report the results on the {FishSeg} test set in Table~\ref{against_baselines}. Our results include 3 statistics, the Intersection-over-Union (IoU) between the predicted foreground mask and the fish true mask, the predicted background mask and the true background mask, and their average (mIoU).

Training on the FishLoc train set, A-LCFCN obtains a significantly higher IoU than LCFCN and PL-FCN methods.
As shown in the qualitative results (Figure~\ref{fig:qualitative}), we see that LCFCN produces small blobs around the center of the objects while PL-FCN outputs large blobs. For both cases, they do not consider the shape of the object as much as A-LCFCN, suggesting that the affinity branch helps in focusing on the segmentation boundaries.

Training on the FishSeg train set which contains less images than FishLoc, the margin improvement between A-LCFCN and LCFCN is smaller. Further, LCFCN performed better when trained on the FishSeg training set than with FishLoc. We observed that the reason behind this result is that LCFCN starts outputting smaller blobs around the object centers the more images it trains on. Thus, it learns to perform better localization at the expense of worse segmentation. On the other hand, A-LCFCN achieved improved segmentation results when trained on the larger training set FishLoc than FishSeg. This result suggests that, with enough images, the affinity branch helps the model focus on achieving better segmentation.

We also report the results of A-LCFCN + PM which shows a consistent improvement over A-LCFCN for both FishLoc and FishSeg benchmarks. This result shows that a fully supervised method can use noisy labels generated from A-LCFCN to further improve the predicted segmentation labels. In Figure~\ref{fig:qualitative} we see that this procedure significantly improves the segmentation boundaries over A-LCFCN's output.

\subsection{Comparison against Full Supervision}

In Table~\ref{against_baselines} we report the results of our methods when fixing the annotation budget. The annotation budget was fixed at around 1500 seconds, which is the estimated it took to annotate the FishLoc dataset. The average time of annotating a single fish and images without fish was one second~\cite{saleh2020realistic}. 

For FS-FCN which was trained on segmentation annotations, the training set consisted of 161 background images and 11 foreground images as it required around 2 minutes to segment a single fish.

We see that A-LCFCN + PM outperforms FS-FCN in this setup by a significant margin, which suggests that with A-LCFCN point-level annotations are more cost-efficient in terms of labeling effort and segmentation performance.

\subsection{Counting and Localization Results}
\begin{table}
\centering
\caption{Counting and Localization Results.}
\label{counting}
\begin{tabular}{lcc}
\toprule
 &   \textbf{MAE} &  \textbf{GAME} \\
\midrule
always-median &   0.575 &  - \\
LCFCN &   0.032 &   0.066 \\
A-LCFCN &   0.057 &   0.066 \\
A-LCFCN+PM &   0.097 &   0.063 \\
\bottomrule
\end{tabular}
\end{table}

To further evaluate the quality of the representations learned by A-LCFCN, we also test it on the FishLoc dataset for the counting and localization tasks. These tasks are essential for marine biologists, which have to assess and track changes in large fish populations~\cite{cui2020fish, Jalal2020FishInformation}. Thus, having a model that automates the localization of these fishes can greatly reduce the cost of tracking large populations, thus helping marine scientist to do efficient monitoring. For our models, the counts are the number of  predicted blobs in the image using the connected components algorithms described in~\citet{laradji2018blobs}.

As a reference, we added the MAE result of `always-median` in Table~\ref{counting} which is a model that outputs a count of $1$ for every test image as it is the median fish count in the training set. We see that although A-LCFCN+PM has improved segmentation over A-LCFCN and LCFCN, the counting and localization counts are very similar. These results \ref{counting} suggest that we can solely use A-LCFCN+PM  for the tasks of segmentation, localization and counting to have a comprehensive analysis of a fish habitat.

\section{Conclusion}
\label{sec:conclusion}
In this paper, we presented a novel affinity-based segmentation method that only requires point-level supervision for efficient monitoring of fisheries. Our approach, A-LCFCN, is trained end-to-end with the LCFCN loss and eliminates the need of explicit supervision for obtaining the pair-wise affinities between pixels. The proposed method combines the output of any standard segmentation architecture with the predicted affinity matrix to improve the segmentation masks with a random walk. Thus, the proposed method is agnostic to the architecture and can be used to improve the segmentation results of any standard backbone. Experimental results demonstrate that A-LCFCN produces significantly better segmentation masks than previous point-level segmentation methods. We also demonstrate that A-LCFCN gets closer to full supervision when used
to generate pseudo-masks to train fully-supervised segmentation network. These results are particularly encouraging for reducing the costs of fish monitoring and achieving sustainable fisheries.

{\small
\bibliography{references}
}

\end{document}